%% file: paper.tex
\pdfoutput=1

\documentclass[sigconf]{acmart}

\usepackage[switch]{lineno}

\usepackage{mathrsfs}
\usepackage{booktabs} 
\usepackage{multicol}
\usepackage{enumitem}
\usepackage{multirow}
\usepackage{hhline}
\usepackage{pifont}
\usepackage{amsmath}
\usepackage{bm}
\usepackage{array}
\usepackage{algorithm}
\usepackage{algpseudocode}
\usepackage{xcolor}
\usepackage{wasysym}
\usepackage{subfigure}
\usepackage{balance}
\usepackage{comment}
\usepackage{amsfonts}

\usepackage{amssymb}

\algnewcommand{\LineComment}[1]{\State \(\#\) #1}

\algnewcommand{\LeftComment}[1]{\State \(\triangleright\) #1}

\newcommand{\rma}{\mathrm{A}}
\newcommand{\rmx}{\mathrm{X}}
\newcommand{\rmd}{\mathrm{d}}
\newcommand{\rmz}{\mathrm{Z}}
\newcommand{\rmza}{\mathrm{Z}^{\mathcal{F}}}
\newcommand{\rmzv}{\mathrm{Z}^{\mathcal{V}}}
\newcommand{\muv}{\mu^{\mathcal{V}}}
\newcommand{\Sigmav}{\Sigma^{\mathcal{V}}}
\newcommand{\mua}{\mu^{\mathcal{F}}}
\newcommand{\Sigmaa}{\Sigma^{\mathcal{F}}}

\newcommand{\underl}{\underline{\mathcal{L}}}
\newcommand{\algo}{\textsc{SeeGera}}

\long\def\comment#1{}

\copyrightyear{2023}
\acmYear{2023}
\setcopyright{acmlicensed}\acmConference[WWW '23]{Proceedings of the ACM Web Conference 2023}{May 1--5, 2023}{Austin, TX, USA}
\acmBooktitle{Proceedings of the ACM Web Conference 2023 (WWW '23), May 1--5, 2023, Austin, TX, USA}
\acmPrice{15.00}
\acmDOI{10.1145/3543507.3583245}
\acmISBN{978-1-4503-9416-1/23/04}

\begin{document}

\title{SeeGera: Self-supervised Semi-implicit Graph Variational Auto-encoders with Masking}



\author{Xiang Li}
\affiliation{%
  \institution{East China Normal University}
  \city{Shanghai}
  \country{China}
}
\email{xiangli@dase.ecnu.edu.cn}

\author{Tiandi Ye}
\affiliation{%
  \institution{East China Normal University}
  \city{Shanghai}
  \country{China}
}
\email{52205903002@stu.ecnu.edu.cn}

\author{Caihua Shan}
\affiliation{%
  \institution{Microsoft Research Asia}
  \city{Shanghai}
  \country{China}
}
\email{caihuashan@microsoft.com}

\author{Dongsheng Li}
\affiliation{%
  \institution{Microsoft Research Asia}
  \city{Shanghai}
  \country{China}
}
\email{dongsheng.li@microsoft.com}

\author{Ming Gao}
\affiliation{%
  \institution{East China Normal University}
  \city{Shanghai}
  \country{China}
}
\email{mgao@dase.ecnu.edu.cn}
\authornote{Corresponding author}






\renewcommand{\shortauthors}{Xiang Li, et al.}

\begin{abstract}
Generative graph self-supervised learning (SSL) aims to learn node representations by reconstructing the input graph data.
However,
most existing methods 
focus on unsupervised learning tasks only
and very few work has shown its superiority over the state-of-the-art graph contrastive learning (GCL) models, especially on the classification task.
While a very recent model has been proposed to bridge the gap,
its performance on unsupervised learning tasks is still unknown.
In this paper,
to comprehensively enhance the performance of generative graph SSL against other GCL models on both unsupervised and supervised learning tasks,
we 
propose the \algo\ model,
which is based on the family of self-supervised variational graph auto-encoder (VGAE).
Specifically,
\algo\ adopts the semi-implicit variational inference framework,
a hierarchical variational framework,
and 
mainly focuses on feature reconstruction and structure/feature masking. 
On the one hand,
\algo\ co-embeds both nodes and features in the encoder and reconstructs both links and features in the decoder.
Since feature embeddings contain rich semantic information on features, 
they can be combined with node embeddings to 
provide fine-grained knowledge for feature reconstruction.
On the other hand,
\algo\ adds an additional layer for structure/feature masking
to the hierarchical variational framework,
which boosts the model generalizability.
We conduct extensive experiments comparing \algo\ with 9 other
state-of-the-art competitors.
Our results show that
\algo\ can compare favorably against other state-of-the-art GCL methods in a variety of unsupervised and supervised learning tasks.
\end{abstract}







\maketitle

\input{tex/introduction.tex}

\input{tex/relatedwork.tex}
\input{tex/preliminary.tex}
\input{tex/algorithm.tex}

\input{tex/experiment.tex}
\input{tex/conclusion.tex}

\section*{Acknowledgement}
This work is supported by Shanghai Pujiang Talent Program No. 21PJ1402900, Shanghai Science and Technology Committee General Program No. 22ZR1419900 and National Natural Science Foundation of China No. 62202172.


\bibliographystyle{ACM-Reference-Format}
\bibliography{paper}

\clearpage 

\appendix

\section{Datasets}
\label{app:data}
We use 7 public datasets which do not have license.
We next briefly introduce them as follows. 

\emph{Cora}, \emph{Citeseer} and 
\emph{Pubmed}~\cite{kipf2016semi}
are three citation networks,
where nodes represent publications and edges are citation links.
Features for each node are the keywords it contains.
Each dimension in the feature vector indicates the presence of a keyword in the publication.
Nodes in these datasets
are associated with labels that describe research topics of publications.

\emph{Coauther CS} and \emph{Coauther Physics} are co-authorship graphs 
based on the Microsoft Academic Graph
from the KDD Cup 2016 challenge~\cite{sinha2015overview}. 
In these datasets,
nodes are authors
and edges capture the co-authorship.
Further,
node features represent keywords in each author’s papers, 
and class
labels indicate the study fields for authors. 

\emph{Amazon Computer} and
\emph{Amazon Photo} are extracted from the Amazon co-purchase graph~\cite{mcauley2015image}, 
where
nodes represent goods and
edges indicate that two goods are frequently bought together.
Node features
are bag-of-words encoded product reviews and class labels are the product categories. 
The statistics of these datasets are summarized in Table~\ref{table:statistics}.

\section{Pseudocodes}
\label{app:pseudo}
This section
summarizes the pseudocodes of \algo-v3 in Alg.~\ref{alg:hoane}.

\begin{algorithm}[H]
\begin{algorithmic}[1]
\Require $\rma$, $\rmx$, $p(\tilde{G}|\rma,\rmx)$,
$\tilde{q}({\epsilon})$, $\hat{q}({\epsilon})$, 
$\rho$, neural networks $T_{\phi_1}$ and $T_{\phi_2}$
\Ensure $\phi_1$ and $\phi_2$
\State Initialize $\phi_1$, $\phi_2$, set $\underl_K^J = 0$
\While{not converged}
\State Sample $\tilde{G} \sim p(\tilde{G}|\rma,\rmx)$
\For{$k = 1$ \text{to} $K$}
\State Sample $\tilde{\psi}_1^k = T_{\phi_1}(\tilde{G}, \tilde{\epsilon}_1^k)$, \text{where} $\tilde{\epsilon}_1^k \sim \tilde{q}({\epsilon})$
\Comment{Eq.~\ref{psi_1}}
\State Sample $\tilde{\psi}_2^k = T_{\phi_2}(\tilde{G}, \tilde{\psi}_1^k, \hat{\epsilon}_2^k)$, \text{where} $\hat{\epsilon}_2^k \sim \hat{q}({\epsilon})$
\Comment{Eq.~\ref{eq:cor}}
\EndFor
\For{$j=1$ \text{to} $J$}
\State Sample ${{\epsilon}}_1^j \sim \tilde{q}({\epsilon})$, ${{\epsilon}}_2^j \sim \hat{q}({\epsilon})$
\State Sample $\psi_1^j = [(\muv)_j, (\Sigmav)_j] = T_{\phi_1}(\tilde{G}, {{\epsilon}}_1^j)$
\Comment{Eq.~\ref{psi_1}}
\State Sample $\psi_2^j = [(\mua)_j, (\Sigmaa)_j] = T_{\phi_2}(\tilde{G}, \psi_1^j, {{\epsilon}}_2^j)$
\Comment{Eq.~\ref{eq:cor}}
\State Sample $\epsilon_j^{\mathcal{V}} \sim \mathcal{N}(0, I)$, $\epsilon_j^{\mathcal{A}} \sim \mathcal{N}(0, I)$ 
\State Sample $(\rmzv)_j = (\muv)_j + (\Sigmav)_j \odot \epsilon_j^{\mathcal{V}}$ 
\State Sample $(\rmza)_j = (\mua)_j + (\Sigmaa)_j \odot \epsilon_j^{\mathcal{A}}$ 
\State  Set $tmp_1 = - \log \Omega_j$  
\Comment{Eq.~\ref{eq:reweight}}
\State  Set $tmp_2 = \log p(\tilde{G} | (\rmzv)_j, (\rmza)_j) $ 
\State  Set $tmp_3 = \log p((\rmzv)_j, (\rmza)_j)$ 
\State Update $\underl_K^J = \underl_K^J + e^{tmp_1 + tmp_2 + tmp_3 } $
\EndFor 
\State Update $\underl_K^J = \log \underl_K^J - \log J$
\State Update $\phi_1 = \phi_1 + \rho \triangledown_{\phi_1} \underl_K^J $ 
\State Update $\phi_2 = \phi_2 + \rho \triangledown_{\phi_2} \underl_K^J $ 
\EndWhile
\State \Return $\phi_1$ and $\phi_2$
\end{algorithmic}
\caption{\algo-v3}
\label{alg:hoane}
\end{algorithm}

\begin{table}[!htbp]
\centering
\caption{Statistics of datasets used in experiments}
\resizebox{0.85\linewidth}{!}
{
\begin{tabular}{c|ccccc} \hline
Datasets &  \#Nodes & \#Edges & \#Features & \#Classes  \\ \hline
Cora  & $2,708$ & $5,278$ & $1,433$ & $7$    \\ 
Citeseer & $3,327$ & $4,676$ & $3,703$ & $6$    \\  
Pubmed & $19,717$ & $88,651$ & $500$ & $3$    \\ 
Coauthor CS & $18,333$ & $327,576$ & $6,805$ & $15$    \\  
Coauthor Physics & $34,493$ & $991,848$ & $8,451$ & $5$    \\  
Amazon Computer & $13,752$ & $574,418$ & $767$ & $10$    \\  
Coauthor Physics & $7,650$ & $287,326$ & $745$ & $8$    \\  \hline
\end{tabular}
}
\label{table:statistics}
\end{table}

\section{Ablation study}
We conduct an ablation study to investigate the main components in \algo.
In particular,
we have extensively compared \algo-v1, \algo-v2 and \algo-v3 in our experiments.
The advantage of \algo-v2 over \algo-v1 shows the importance of capturing the correlations between node and feature embeddings.
Also, 
the outperformance of \algo-v3 over \algo-v2 verifies the importance of the masking mechanism.
Further,
to show the effectiveness of our proposed feature reconstruction method,
we remove feature embeddings in the encoder and feed only node embeddings into GCN in the decoder to reconstruct features.
We call this variant \algo\_nf (\textbf{n}o \textbf{f}eature embedding).
Table~\ref{results:ai:ab}
shows the results on attribute inference.
We exclude \algo-v3 in the table, because it further uses the masking mechanism while others not.
From the table,
we see that both \algo-v1 and \algo-v2 outperform \algo\_nf.
This shows the importance of using both node and feature embeddings for feature reconstruction.

\begin{table}[!htbp]
\centering
\caption{The comparison between \algo\ and \algo\_{nf} in the attribute inference task.
}
\resizebox{\linewidth}{!}
{
\begin{tabular}{c|ccccc} \hline
Method & Cora & Citeseer & CS \\ 
\hline 
\algo\_{nf} & $1.91 \times 10^{-3} \pm 3.78 \times 10^{-5}$ & $5.15 \times 10^{-4} \pm 1.02 \times 10^{-6}$ & $2.14 \pm 0.07$ \\
\algo-v1& $1.90 \times 10^{-3} \pm 8.18 \times 10^{-5}$ & $4.87 \times 10^{-4} \pm 2.62 \times 10^{-6}$ & $2.12 \pm 0.06$ \\ 
\algo-v2& $\bm{1.89 \times 10^{-3} \pm 8.13 \times 10^{-5}}$ & $\bm{4.86 \times 10^{-4} \pm 2.24 \times 10^{-6}}$ & $\bm{2.08 \pm 0.07}$ \\
\hline
\end{tabular}
}
\label{results:ai:ab}
\end{table}

\section{Implementation Details}
\label{app:imp}
We implemented \algo\ by PyTorch.
The model is initialized by Glorot initialization and trained by Adam.
We run the model for 3500 epochs 
on all the datasets. 
In particular,
for the task of node classification, we further run the logistic regression classifier for 500 epochs. 
We adopt GCNs in both encoder and decoder for all the datasets except CS.
For CS,
we use MLPs to replace GNNs instead.
We choose Bernoulli noise \texttt{Ber}(0.5) for both $\tilde{\epsilon} \sim \tilde{q}({\epsilon})$ and $\hat{\epsilon} \sim \hat{q}({\epsilon})$,
and
set the noise dimension to 5 in all tasks. 
As suggested by~\cite{sonderby2016ladder},
we use 
warm-up during the first 300 epochs 
to gradually impose the {prior regularization terms} $\mathrm{D_{KL}} (q_{1}(\rmzv | \psi_1) || p(\rmzv))$
and $\mathrm{D_{KL}} (q_{2}(\rmza | \psi_2) || p(\rmza))$.
For both priors $p(\rmzv)$ and $p(\rmza)$,
we assume that they follow standard multivariate normal distributions.
In the node classification task,
most results of baselines 
are publicly available and we directly report these results from their original papers.
For the results that are missing,
we run the released codes by their authors and fine-tune the model hyper-parameters.
In the link prediction and attribute inference tasks,
since most baselines are not studied in these tasks,
we run their released codes with fine-tuning.
For fairness,
we 
run all the experiments on a server with a single NVIDIA A100 GPU with 80G memory.
For simplicity,
we set $J = K = 1$ in Eq.~\ref{eq:reweight} for both \algo\ and SIG-VAE.
One may further refer to~\cite{rainforth2018tighter} for better value selection.
{
All the datasets and codes are provided at \url{https://github.com/SeeGera/SeeGera}.
}
We provide the detailed hyper-parameter settings of \algo-v3 on different datasets in Tables~\ref{hyper-parameters:lp}-~\ref{hyper-parameters:cla}. All hyper-parameters are selected through small grid search, and the search space is provided as:
\begin{itemize}
    \item Number of layers in the encoder $L_1$: \{1, 2, 3\}
    \item Number of layers in the decoder $L_2$: \{1, 2, 3\}
    \item Learning rate of \algo: \{1e-3, 5e-3, 1e-2\}
    \item Dropout of \algo: \{0, 0.1, 0.3, 0.5, 0.7, 0.9\}
    \item Weight decay of \algo: \{5e-5, 1e-4, 5e-4, 1e-3\}
    \item Structure masking rate $\alpha_1$: \{0, 0.1, 0.2, 0.3, 0.4, 0.5\}
    \item Feature masking rate $\alpha_2$: \{0, 0.1, 0.2, 0.3, 0.4, 0.5\}
    \item Learning rate of logistic regression: \{1e-3, 5e-3, 1e-2\}
    \item Dropout of logistic regression: \{0, 0.1, 0.3, 0.5, 0.7, 0.9\}
    \item Weight decay of logistic regression: \{5e-5, 1e-4, 5e-4, 1e-3\}
\end{itemize}

\begin{table}[!htbp]
    \centering
    \caption{Hyper-parameter setting details of \algo-v3 in link prediction.}
    \resizebox{0.85\linewidth}{!}
{
    \begin{tabular}{cccccccc}
    \toprule
    Dataset     & $L_1$ & $L_2$ & lr    & dropout   & wd    & $\alpha_1$    & $\alpha_2$ \\ \hline
    Cora        & 2     & 3     & 1e-3  & 0.3       & 5e-5  & 0.3           & 0.0 \\
    Citeseer    & 1     & 2     & 1e-3  & 0.0       & 1e-4  & 0.6           & 0.0 \\
    Pubmed      & 2     & 1     & 5e-3  & 0.5       & 0.0   & 0.3           & 0.1 \\
    Photo       & 1     & 1     & 5e-3  & 0.0       & 0.0   & 0.5           & 0.0 \\
    Computer    & 2     & 2     & 1e-3  & 0.0       & 0.0   & 0.4           & 0.0 \\
    CS          & 1     & 2     & 1e-3  & 0.0       & 0.0   & 0.5           & 0.5 \\
    Physics     & 2     & 2     & 1e-3  & 0.0       & 0.0   & 0.2           & 0.0 \\
    \bottomrule
    \end{tabular}
    }
    \label{hyper-parameters:lp}
\end{table}
\begin{table}[!htbp]
    \centering
    \caption{Hyper-parameter setting details of \algo-v3 in attribute inference.}
    \resizebox{0.85\linewidth}{!}
{
    \begin{tabular}{cccccccc}
    \toprule
    Dataset     & $L_1$ & $L_2$ & lr    & dropout   & wd    & $\alpha_1$    & $\alpha_2$ \\ \hline
    Cora        & 1     & 1     & 1e-3  & 0.1       & 5e-5  & 0.4           & 0.0 \\
    Citeseer    & 1     & 1     & 1e-3  & 0.0       & 0.0   & 0.3           & 0.0 \\
    Pubmed      & 2     & 2     & 1e-3  & 0.0       & 0.0   & 0.0           & 0.1 \\
    Photo       & 2     & 3     & 1e-3  & 0.0       & 0.0   & 0.3           & 0.0 \\
    Computer    & 2     & 3     & 1e-3  & 0.0       & 0.0   & 0.2           & 0.0 \\
    CS          & 1     & 3     & 1e-3  & 0.0       & 0.0   & 0.1           & 0.5 \\
    Physics     & 2     & 2     & 1e-3  & 0.0       & 0.0   & 0.0           & 0.0 \\
    \bottomrule
    \end{tabular}
    }
    \label{hyper-parameters:ai}
\end{table}
\begin{table}[!htbp]
    \centering
    \caption{Hyper-parameter setting details of \algo-v3 in node classification.}
    \resizebox{\linewidth}{!}
{
    \begin{tabular}{cccccccc|ccc}
    \toprule
        \multirow{2}{*}{Dataset} 
        & \multicolumn{7}{c|}{\algo} & \multicolumn{3}{c}{Logistic Regression} \\ \cline{2-11}
                    & $L_1$ & $L_2$ & lr    & dropout   & wd    & $\alpha_1$    & $\alpha_2$    & lr    & dropout   & wd \\ \hline
        Cora        & 2     & 2     & 1e-3  & 0.3       & 1e-3  & 0.2           & 0.1           & 1e-3  & 0.9       & 1e-3 \\
        Citeseer    & 2     & 2     & 5e-3  & 0.3       & 5e-4  & 0.0           & 0.0           & 1e-3  & 0.9       & 0.0 \\
        Pubmed      & 2     & 1     & 5e-3  & 0.7       & 0.0   & 0.0           & 0.0           & 1e-3  & 0.3       & 0.0 \\
        Photo       & 1     & 2     & 1e-3  & 0.0       & 5e-4  & 0.5           & 0.5           & 5e-3  & 0.7       & 5e-5 \\
        Computer    & 1     & 3     & 5e-3  & 0.0       & 5e-4  & 0.0           & 0.0           & 5e-3  & 0.3       & 1e-4 \\
        CS          & 2     & 3     & 1e-2  & 0.1       & 1e-3  & 0.4           & 0.0           & 1e-2  & 0.0       & 1e-3 \\
        Physics     & 1     & 3     & 1e-2  & 0.0       & 0.0   & 0.0           & 0.0           & 1e-2  & 0.5       & 5e-5 \\
    \bottomrule
    \end{tabular}
    }
    \label{hyper-parameters:cla}
\end{table}

\section{Variational lower bound}
\label{app:lb}
In this section,
we show the derivation on the variational lower bounds in detail.
\begin{equation}
\label{eq:underll}
\nonumber
\small
\begin{split}
 \mathcal{L} & = \mathbb{E}_{h_{\phi_1}(\rmzv)} \mathbb{E}_{h_{\phi_2}(\rmza)} \left [ \log \frac{p (\rmzv | \rma,\rmx) p(\rmza| \rmx^T)p(\rma,\rmx)}{h_{\phi_1}(\rmzv) h_{\phi_2}(\rmza)} \right ]\\ 
& = -\mathrm{D_{KL}} (h_{\phi_1}(\rmzv) || p (\rmzv | \rma, \rmx)) -\mathrm{D_{KL}} (h_{\phi_2}(\rmza) || p (\rmza | \rmx ^ T))+ \log p(\rma,\rmx)\\ 
& \geq -\mathbb{E}_{\psi_1 \sim q_{\phi_1}(\psi_1)} \mathrm{D_{KL}} (q_{1}(\rmzv | \psi_1) || p (\rmzv | \rma, \rmx)) \\
& \quad \quad - \mathbb{E}_{\psi_2 \sim q_{\phi_2}(\psi_2)} \mathrm{D_{KL}} (q_{2}(\rmza | \psi_2) || p (\rmza | \rmx^T)) + \log p(\rma,\rmx) \\
& = \mathbb{E}_{\psi_1 \sim q_{\phi_1}(\psi)}\mathbb{E}_{\rmzv \sim q_{1}(\rmzv | \psi_1)}\mathbb{E}_{\psi_2 \sim q_{\phi_2}(\psi)}\mathbb{E}_{\rmza \sim q_{2}(\rmza | \psi_2)}\\
& \quad \quad \left [ \log \frac{p (\rma,\rmx,\rmzv,\rmza)}{q_{1}(\rmzv | \psi_1)q_{2}(\rmza | \psi_2)}\right ] \\
& = \underl_1, \\
\end{split}
\end{equation}
where $\mathrm{D_{KL}}$ is the KL divergence
and
we employ 
$\mathrm{D_{KL}}(\mathbb{E}_{\psi}q(\rmz | \psi) || p(\rmz)) \leq \mathbb{E}_{\psi} \mathrm{D_{KL}}(q(\rmz | \psi ) || p(\rmz))$
according to~\cite{pmlr-v80-yin18b}.
To better understand $\underl_1$,
we decompose the joint distribution $p (\rma,\rmx,$ $\rmzv,\rmza)$ as 
\begin{equation}
\nonumber
\small
\begin{split}
& p (\rma,\rmx,\rmzv,\rmza)  = p(\rmzv) p(\rmza) \prod_{i,j \in \mathcal{V}} p(\rma_{ij}| \rmzv_i, \rmzv_j) \prod_{i \in \mathcal{V}, r \in \mathcal{F}}p(\rmx_{ir} | \rmzv_i, \rmza_r)
\end{split}
\end{equation}
and expand $\underl_1$ to derive:
\begin{equation}
\nonumber
\begin{split}
\underl_1 & =  \mathbb{E}_{\psi_1 \sim q_{\phi_1}(\psi_1)}\mathbb{E}_{\rmzv \sim q_{1}(\rmzv | \psi_1)} \left [\sum_{i,j \in \mathcal{V}} \log p(\rma_{ij}| \rmzv_i, \rmzv_j) \right ]\\ 
&\quad + \mathbb{E}_{\psi_1 \sim q_{\phi_1}(\psi_1)}\mathbb{E}_{\rmzv \sim q_{1}(\rmzv | \psi_1)}\mathbb{E}_{\psi_2 \sim q_{\phi_2}(\psi_2)}\mathbb{E}_{\rmza \sim q_{2}(\rmza | \psi_2)} \\
& \quad \quad \left [\sum_{i \in \mathcal{V}, r \in \mathcal{F}} \log p(\rmx_{ir}| \rmzv_i, \rmza_r) \right ]\\ 
&\quad - \mathbb{E}_{\psi_1 \sim q_{\phi_1}(\psi_1)} \mathrm{D_{KL}} (q_{1}(\rmzv | \psi_1) || p(\rmzv)) \\
& \quad - \mathbb{E}_{\psi_2 \sim q_{\phi_2}(\psi_2)} \mathrm{D_{KL}} (q_{2}(\rmza | \psi_2) || p(\rmza)).
\end{split}
\end{equation}
Here,
$q_{1}(\rmzv | \psi_1)$ and $q_{2}(\rmza | \psi_2)$ 
are encoders that
generate embeddings of nodes and features, respectively;
$p(\rma_{ij}| \rmzv_i, \rmzv_j)$ and $p(\rmx_{ir}| \rmzv_i, \rmza_r)$
are decoders 
that reconstruct links and features from learned embeddings.
The first two terms in the equation
correspond to the negative reconstruction loss for links and features,
while the last two terms are regularizers that promote the closeness between variational distributions and prior distributions.

Similarly,
we can expand $\underl_2$ as:
\begin{equation}
\nonumber
\begin{split}
\underl_2 & = \mathbb{E}_{\psi_1 \sim q_{\phi_1}(\psi_1)}\mathbb{E}_{\psi_2 \sim q_{\phi_2}(\psi_2 \vert \psi_1)}\mathbb{E}_{(\rmzv,\rmza) \sim q(\rmzv,\rmza | \psi_1,\psi_2)}\\
& \quad \quad \left [\sum_{i,j \in \mathcal{V}} \log p(\rma_{ij}| \rmzv_i, \rmzv_j) + \sum_{i \in \mathcal{V}, r \in \mathcal{F}} \log p(\rmx_{ir}| \rmzv_i, \rmza_r) \right ]\\ 
&\quad - \mathbb{E}_{(\psi_1,\psi_2) \sim q_{\phi}(\psi_1,\psi_2)} \mathrm{D_{KL}} (q(\rmzv,\rmza | \psi_1,\psi_2) || p(\rmzv,\rmza)).
\end{split}
\end{equation}

\end{document}

%% file: tex/introduction.tex
\section{Introduction}
\label{sec:intro}
Self-supervised learning (SSL)~\cite{he2020momentum,devlin2018bert,you2020does,hu2020gpt} has attracted significant attention recently. 
By extracting and 
employing supervisions from data itself,
SSL can heavily reduce the dependence 
of neural network models
on the labeled data, 
which is costly to obtain.
To facilitate 
graph-based learning,
SSL has 
been applied on graph-structured data.
For example,
it can
learn representations for nodes (e.g., web pages in search engines), and detect the anomalies on webs (e.g., malicious users)~\cite{liu2021anomaly}. 
Recently,
graph contrastive learning (GCL),
as one of the main SSL types, 
has experienced a surge~\cite{velickovic2019deep,sun2019infograph,you2020graph,jiao2020sub}.
The core idea of GCL is 
to first construct positive and negative pairs for nodes, 
and then maximize the similarity between positive pairs while minimizing that between negative ones\footnote{Note that some GCL methods require positive pairs only and they only maximize the similarity between positive pairs.}.


Despite the success, 
existing GCL methods suffer from two main problems.
On the one hand,
negative samples are needed in most contrastive objectives, 
which generally construct one positive and $K$ negative samples for each node.
However, these models are easily affected by the value of $K$.
When $K$ is small,
the model cannot learn sufficient discriminative information, 
which degrades the model effectiveness;
otherwise,
there could lead to a large number of
false-negative samples and slow convergence.
Generally,
$K$ is set empirically
and there lack theoretical supports.
On the other hand,
for the rest of methods based on positive pairs only,
they are easily trapped into a degenerate solution~\cite{zhu2021empirical}, 
where all the output embeddings of nodes collapse to
a constant. 
To tackle the issue, 
additional strategies are necessary, such as asymmetric dual encoders
with momentum updates and exponential
moving average~\cite{qiu2020gcc,thakoor2021large}.
Recently,
some studies~\cite{li2022understanding} have showed that although these training strategies can alleviate collapse to some extent, they may still cause collapse in partial dimensions of the representation, which leads to worse performance.

To address the shortcomings of GCL methods,
generative graph SSL methods can be used instead.
In particular,
self-supervised 
graph auto-encoders (GAEs)~\cite{kipf2016variational},
whose objective is to reconstruct the input graph data,
have been widely studied.
Existing methods mainly differ in their adopted reconstruction components,
such as the adjacency matrix reconstruction~\cite{pan2018adversarially}, 
the node feature reconstruction~\cite{park2019symmetric} 
and a combination of both graph structure and node feature reconstruction~\cite{salehi2019graph}.
However, most of these methods 
focus on the unsupervised learning tasks like link prediction and node clustering,
and very few work has shown its superiority over the state-of-the-art GCL methods,
especially on the classification task.
While
a masked GAE model GraphMAE~\cite{hou2022graphmae} is very recently proposed to bridge the gap,
its performance on the unsupervised learning tasks is still unexplored.
Since the goal of SSL is to learn versatile representations,
a further study on self-supervised GAE model
that can achieve comprehensive superiority
on both unsupervised and supervised learning tasks is needed.
Further,
although GraphMAE
is an auto-encoding method,
it is based on GAE and is essentially not a generative model.
This also calls our attention back to the study of generative graph SSL model,
such as variational graph auto-encoder (VGAE)~\cite{kipf2016variational}.


Different from GAE,
VGAE
consists of
an inference model and a generative model.
Specifically,
the inference model encodes observations (links and features) into latent variables (node embeddings)
while the generative model decodes from these latent variables to reconstruct links.
However,
as pointed out in~\cite{hou2022graphmae},
node feature reconstruction is 
beneficial for learning high-quality representations.
Therefore,
the lack of feature reconstruction could degrade the model effectiveness.
To solve the issue,
most
existing methods 
adopt MLP~\cite{hu2019strategies,hu2020gpt} and GNNs~\cite{hou2022graphmae,park2019symmetric} 
as their decoders
for feature reconstruction.
However,
they utilize node-level embeddings only
and
ignore feature-level embeddings that contain rich semantic information on node features and can be used to help feature reconstruction.
Recently,
CAN~\cite{meng2019co}
is proposed to co-embed both nodes and features,
and use the inner product of their embeddings as the decoder to recover node features.
Despite the success,
it has three main problems.
First,
the linear decoder is generally less powerful than MLP and GNNs, 
which restricts the model's capability in reconstructing node features.
Second,
it assumes the independence between node and feature embeddings in the variational inference stage, but practically these two types of embeddings are highly correlated.
Third,
it lacks structure/feature masking in the learning process,
which has been shown to degrade the model's performance on the classification task~\cite{hou2022graphmae}.

In this paper,
we study generative graph SSL and 
our goal is to enhance the family of self-supervised VGAE on graph representation learning 
in a variety of downstream tasks.
Recently,
semi-implicit variational inference (SIVI)~\cite{pmlr-v80-yin18b},
which is a hierarchical variational framework,
has been applied to VGAE to 
model a wide range of underlying true posteriors with multi-modality, skewness and heavy tails~\cite{hasanzadeh2019semi}.
We thus adopt the framework to remove the explicit Gaussian restriction on the variational distribution and 
mainly focus on the component of feature reconstruction and structure/feature masking.
We
propose
a \textbf{Se}lf-supervised s\textbf{e}mi-implicit \textbf{G}raph variational auto-encod\textbf{er}
with m\textbf{a}sking,
namely, 
{\algo}.
Specifically,
the model 
co-embeds both nodes and features in the
encoder and
jointly reconstructs links and features in the 
decoder.
Note that
the feature embeddings can provide fine-grained information
that is supplementary to the node embeddings 
when reconstructing node features.
Specifically,
for each node,
we take its feature values as weights and compute the weighted average of feature embeddings w.r.t. the node.
The weighted embedding characterizes the affinities between the node and all the features.
After that,
we 
combine the weighted embedding with the node embedding,
and feed the fused embedding into GNNs to reconstruct the node's features.
Further, 
to generate node and feature embeddings in the encoder,
we first assume the independence between them
and propose the base \algo\ model.
Then
we upgrade the model by capturing the correlations between node and feature embeddings.
Finally,
we add an additional layer to the hierarchical variational framework to integrate \algo\ with the masking mechanism and boost the model performance.
In summary,
our main contributions are listed:

\noindent{\small$\bullet$}
We propose a generative graph SSL model \algo.
To our knowledge,
this is the first generative graph SSL method that is comprehensively compared with the SOTA GCL models in terms of both unsupervised and supervised learning tasks,
and shows superiority.

\noindent{\small$\bullet$}
We present a novel feature reconstruction method that 
leverages both node and feature embeddings to provide fine-grained information for reconstructing features.
We further introduce the structure/feature masking mechanism by adding an additional layer to the hierarchical variational framework.

\noindent{\small$\bullet$}
We conduct extensive experiments to evaluate the performance of \algo\ 
on two unsupervised learning tasks: link prediction and attribute inference,
and one supervised learning task: node classification. 
Experimental results show that
\algo\ can significantly outperform other competitors on both link prediction and attribute inference tasks,
and perform comparably with them in node classification.
This effectively verifies the power of generative graph SSL in graph representation learning.

\comment{
Data analytics on HINs has been an active area of research~\cite{sun2011pathsim,li2016transductive}. 
Being a fundamental task in machine learning and data mining, cluster analysis has found interesting 
applications in HINs.
For example, clustering Facebook users based on their interests enables effective target and viral marketing~\cite{li2017semi}.
Even though spectral clustering is very effective for data that is modeled as
(homogeneous) network/graph~\cite{liu2013large}, there are surprisingly few studies that apply spectral clustering to HINs.
The objective of this paper is to study how spectral clustering can be effectively applied to HINs to improve clustering quality. 
%

Spectral clustering
transforms clustering into a graph partitioning problem
that optimizes a certain criterion that measures the quality of the partitions, 
such as the \emph{normalized cuts}~\cite{shi2000normalized}.
Generally, given a set of objects $X = \{x_1,x_2,..., x_n\}$,
standard spectral clustering methods
first construct an undirected graph $G = (X, S)$,
where $X$ denotes the vertex set and 
$S$ is a matrix such that $\sij$ 
measures the similarity between objects $x_i$ and $x_j$.\footnote{Given a matrix $M$, we use $M_{ij}$ or $M[i,j]$ to refer to the ($i$,$j$)-th entry of $M$.}
Then, the Laplacian matrix $L_S$ is computed based on which eigen-decomposition
is performed to obtain $k$ eigenvectors that correspond to the $k$ smallest eigenvalues,
where $k$ is the number of desired clusters.
These eigenvectors are used as new feature space of objects.
Finally, a post-processing step, such as $k$-means~\cite{ng2002spectral}
and spectral rotation~\cite{stella2003multiclass} is applied to partition the objects into $k$ clusters.

Previous studies have shown that the performance of spectral clustering highly depends on the
``quality" of the similarity matrix~\cite{nie2016constrained}.
Intuitively, a high-quality matrix $S$ is one 
such that $\sij$ is large if objects $x_i$ and $x_j$ ought to be in the same cluster and
$\sij$ is small otherwise.
The challenges of constructing a high-quality matrix $S$ in HINs are two-fold.
First, although the similarity between two objects in an HIN can be measured
by conventional network distances (such as shortest paths or random-walk based similarity), 
previous works have shown that \emph{meta-path/meta-structure based similarity}
is much more effective in HINs~\cite{sun2011pathsim,huang2016meta,fang2016semantic}.
(We use meta-paths in this paper and our method can be easily adapted to using meta-structures.)
A meta-path is a sequence of object types that expresses a 
path-based relation between two objects.
For example, in Facebook,
the meta-path \emph{User-Event-User} represents the relation between users who have attended the same event; the meta-path
\emph{User-Page-User} captures the relation between two users who have \emph{liked} the same product page.
An interesting issue is how various meta-paths can be integrated to formulate a similarity matrix 
that exhibits a clear clustering structure.
Second,
a theoretically infinite number of meta-paths can be derived from an HIN (with meta-paths composed of different object types and of various lengths).
However, generally, 
only a small subset of them are useful for a given clustering task.
For example,
the meta-path \emph{User-Page-User} is useful for clustering users based on their interests.
The same meta-path, however, is much less useful if we want to cluster users based on their geographic locations.
A mechanism for weighing the relative importance of meta-paths is thus essential.
%
\comment{
Third, 
spectral clustering adopts a post-processing step to return clusters.
However,
different post-processing steps may lead to 
different clustering results
and there does not exist the optimal one~\cite{huang2013spectral}.
Given a graph with $k$ (the number of clusters) independently connected components,
each component will be identified as a cluster~\cite{meila2001random}.
In this case, the post-processing step can be removed.
Therefore, constructing a graph with very clear cluster structure is much preferred.
}

We propose the \algo\ algorithm, which stands for \underline{S}pectral \underline{Cl}ustering \underline{U}sing \underline{M}eta-\underline{P}aths, to address the above problems. 
\algo\ uses meta-paths to construct the similarity matrix $S$. The matrix $S$ is  refined through an iterative process, whose goal is to optimize an objective function that captures the quality of $S$.
During the process, weights of meta-paths are also learned. 
Here, we summarize our contributions.

\noindent{\small$\bullet$}
We show how spectral clustering can be effectively applied to HINs.
In particular, we show how meta-paths are used to construct an effective similarity matrix.

\noindent{\small$\bullet$}
We propose the \algo\ algorithm, which employs an iterative learning process via which 
the similarity matrix and weights of meta-paths
are mutually refined.

\noindent{\small$\bullet$}
We conduct extensive experiments on real datasets to show the effectiveness of \algo.
We compare \algo\ with the state-of-the-art clustering methods for HINs.
Our results show that spectral clustering is an effective approach for HINs and that \algo\ 
significantly outperforms existing methods.

The rest of the paper is organized as follows.
Section~\ref{sec:relatedwork}
summarizes related works on  general spectral clustering and existing clustering methods for HINs.
Section~\ref{sec:definition}
gives formal definitions of related concepts and the problem we study.
Section~\ref{sec:algorithm} describes our algorithm \algo.
Section~\ref{sec:exp} gives the experimental results and 
Section~\ref{sec:conclusion} concludes the paper.
}

%% file: tex/relatedwork.tex
\section{Related Work}
\label{sec:relatedwork}
In this section,
we summarize the related work on both graph self-supervised learning and 
generative graph self-supervised learning,
respectively.

\subsection{Graph self-supervised learning}
Graph self-supervised learning~\cite{sun2019infograph,you2020graph,hou2022graphmae,xu2021self}
aims to employ supervisions extracted from graph-structured data
without the need for annotated data.
Existing methods can be mainly divided into four types:
(1) generative models~\cite{kipf2016variational},
whose objective is to reconstruct the input graph data.
(2) auxiliary-property-based methods~\cite{you2020does}, which first obtain graph-related properties and then take them as supervisions,
such as the pseudo labels of unlabeled nodes; 
(3) contrastive models~\cite{velickovic2019deep}, 
which construct positive and negative pairs for contrast.
(4) hybrid approaches~\cite{zhang2020graph}, which
combine the objectives of the first three types in a multi-task learning fashion. 
For a comprehensive survey on
graph self-supervised learning, see~\cite{liu2022graph}.

Recently,
graph contrastive learning 
has been widely studied.
According to whether negative samples are used in the learning process,
existing methods include negative-sample-based and negative-sample-free ones.
For the former,
DGI~\cite{velickovic2019deep} and InfoGraph~\cite{sun2019infograph} employ corruptions to construct negative pairs.
GRACE~\cite{zhu2020deep}, GCA~\cite{zhu2021graph} and GraphCL~\cite{you2020graph}
take
samples in a mini-batch as a dictionary whose size is constrained by the batch size
and consider 
other samples in the same mini-batch as negatives of a sample,
while 
GCC~\cite{qiu2020gcc} maintains a dynamic dictionary with larger size as in MoCo~\cite{he2020momentum}.
For the latter,
BGRL~\cite{thakoor2021large} 
and 
CCA-SSG~\cite{zhang2021canonical} are two representative models that are based on asymmetric encoding architectures.
However, 
they require special
training strategies to avoid 
the collapse of learned node embeddings to a constant, such as
momentum update~\cite{he2020momentum},
exponential moving average~\cite{thakoor2021large} and
stop gradient~\cite{thakoor2021large}.
Further,
existing GCL methods heavily rely on graph augmentation strategies 
to construct different graph views 
for contrast,
including feature-oriented (e.g., masking~\cite{you2020graph} and shuffling~\cite{velickovic2019deep}), proximity-oriented (e.g., perturbation~\cite{you2020graph}), and graph-sampling-based (e.g., random-walk~\cite{hassani2020contrastive}) augmentations.


\subsection{Generative graph self-supervised learning}
Generative graph self-supervised learning aims to take the input graph as self-supervision and recover the input data.
It mainly consists of 
two families of models:
graph autoregressive models and graph autoencoders (GAEs).
Autoregressive models~\cite{you2018graphrnn,you2018graph} decompose joint probability distributions as a product of conditionals.
The representative graph autoregressive model is GPT-GNN~\cite{hu2020gpt},
which takes attributed graph generation as its objective.
However, since autoregressive models require an explicit ordering to generate,
they might not work well on
graphs that do not exhibit inherent orders.

Different from graph autoregressive models,
GAEs do not require any decoding ordering
and they aim to reconstruct part of the input graph data.
According to the reconstructed components,
existing self-supervised GAE methods include those that reconstruct links only (e.g., ARVGA~\cite{pan2018adversarially}, GAE~\cite{kipf2016variational}, VGAE~\cite{kipf2016variational}),
features only (e.g., GraphMAE~\cite{hou2022graphmae}, GALA~\cite{park2019symmetric},
MGAE~\cite{wang2017mgae}, EP~\cite{garcia2017learning}),
and a combination of both links and features (e.g., GATE~\cite{salehi2019graph}, CAN~\cite{meng2019co},
DGE~\cite{zhoudge}).
However,
most of these methods
focus on the link prediction and node clustering tasks,
and few of them compares favorably against the state-of-the-art GCL methods, especially in the classification task.
While GraphMAE is recently proposed to bridge the gap,
its performance on unsupervised learning tasks remains unexplored. Further, 
it is based on GAE and is essentially not a generative model.
Different from GAE,
variational graph auto-encoder (VGAE) is a generative model
that recovers links only in the decoder.
While there exist some self-supervised VGAE models that 
reconstruct features~\cite{salehi2019graph,zhoudge},
most of them 
only leverage node-level embeddings but ignore feature-level embeddings that contain fine-grained information for node features and can help boost feature reconstruction.
In this paper,
we reconsider generative graph self-supervised learning
and show that 
self-supervised VGAE
can outperform or perform comparably against 
other SOTA GCL models
in a variety of tasks, 
such as link prediction, attribute inference and node classification. 

%% file: tex/preliminary.tex
\section{Preliminary}
\label{sec:bg}

\subsection{Notations}
Let $\mathcal{G} = (\mathcal{V}, \mathcal{E})$ denote a graph,
where $\mathcal{V} = \{x_i\}_{i=1}^n$ is a set of nodes and 
$\mathcal{E} \subseteq \mathcal{V} \times \mathcal{V}$ is a set of edges.
Let 
$\mathrm{A}$ 
be 
the adjacency matrix of $G$,
such that $\rma_{ij}$ represents the weight of edge $e_{ij}$ between objects $x_i$ and $x_j$.
For simplicity,
we set $\rma_{ij} = 1$ if $e_{ij} \in \mathcal{E}$; $0$, otherwise.
Further,
since nodes in a graph are usually associated with features,
we denote 
$\mathcal{F} = \{f_r\}_{r=1}^l$ as a set of node features
and $\mathrm{X} \in \mathbb{R}^{n\times l}$ as the node feature matrix,
where the $i$-th row $\rmx_i$ is the feature vector of node $x_i$.
For the node representation matrix,
let it be $\rmzv \in \mathbb{R}^{n\times d}$,
where $d$ is the output embedding dimension satisfying $d \ll \vert \mathcal{V} \vert$.
Note that the $i$-th row $\rmzv_i$ represents the embedding of node $x_i$.
Similarly,
$\rmza \in \mathbb{R}^{l\times d}$ denotes the feature representation matrix,
whose $r$-th row $\rmza_r$ is the embedding of node feature $f_r$.
In this paper,
we learn both node and feature representations, and use node representations in various downstream tasks.

\subsection{SIVI and SIG-VAE}
Given observations $\textrm{Y}$ and latent variable $\rmz$,
the vanilla variational inference (VI) derives an evidence lower bound 
\begin{equation}
    \text{ELBO} =  - \mathbb{E}_{ \rmz \sim q( \rmz | \psi)} \left [\log q(\rmz|\psi) - \log p(\textrm{Y},\rmz) \right ],
\end{equation}
where 
$\psi$ is variational parameter,
$q(\rmz | \psi)$ is variational distribution and $p(\textrm{Y}, \rmz)$ is joint distribution.
However, VI restricts an exponential family assumption to the posterior.
To address the problem,
semi-implicit variational inference (SIVI)~\cite{pmlr-v80-yin18b}
considers variational parameters as random variables drawn from a mixing distribution.
Specifically, 
the semi-implicit variational distribution for 
$\rmz$
is defined in a hierarchical manner, which follows
$\rmz \sim q(\rmz \vert \psi)$ and $\psi \sim q_{\phi}(\psi)$.
Here,
$\phi$ is the parameter of the mixing distribution $q_{\phi}(\psi)$.
Further,
$\psi$ can be marginalized out to derive a distribution family $\mathcal{H}$ indexed by $\phi$ for $\rmz$:
\begin{equation}
\mathcal{H} = \left\{ h(\rmz): h(\rmz) = \int_{\psi}q(\rmz \vert \psi)q_{\phi}(\psi)\rmd \psi \right\}.
\end{equation}
Note that $q(\rmz \vert \psi)$ is required to be explicit,
but the mixing distribution $q_{\phi}(\psi)$ is allowed to be implicit.
Moreover,
the marginal distribution $h(\rmz) \in \mathcal{H}$ is often implicit unless 
$q_{\phi}(\psi)$ is conjugate to $q(\rmz \vert \psi)$.
These are the reasons why the method is referred to as ``semi-implicit'' VI.
To maintain simple optimization,
$q(\rmz \vert \psi)$ is required 
to be either reparameterizable~\cite{kingma2013auto}
or allow the ELBO under $q(\rmz \vert \psi)$ to be analytic.
For $q_{\phi}(\psi)$,
it needs to be reparameterizable.
Generally,
SIVI draws from $q_{\phi}(\psi)$ by injecting random noise $\epsilon$ into node features and transforming the features via neural networks.

Recently, \citet{hasanzadeh2019semi} apply
SIVI to VGAE and propose the semi-implicit graph variational auto-encoder (SIG-VAE) model.
Specifically,
it sets $q(\rmz \vert \psi)$ to be Gaussian distribution and 
uses GNNs to characterize the mixing distribution 
$q_{\phi}(\psi)$.
While SIG-VAE uses the hierarchical variational framework to capture complex non-Gaussian posteriors,
it still has the problem of ignorance of feature reconstruction
and structure/feature masking.
Therefore, 
based on the framework of SIG-VAE,
we next explore how to enhance self-supervised VGAE for unsupervised graph representation learning.

%% file: tex/algorithm.tex
\section{Algorithm}
\label{sec:algorithm}
\begin{figure*}[t]
  \centering
  \includegraphics[width=\textwidth]{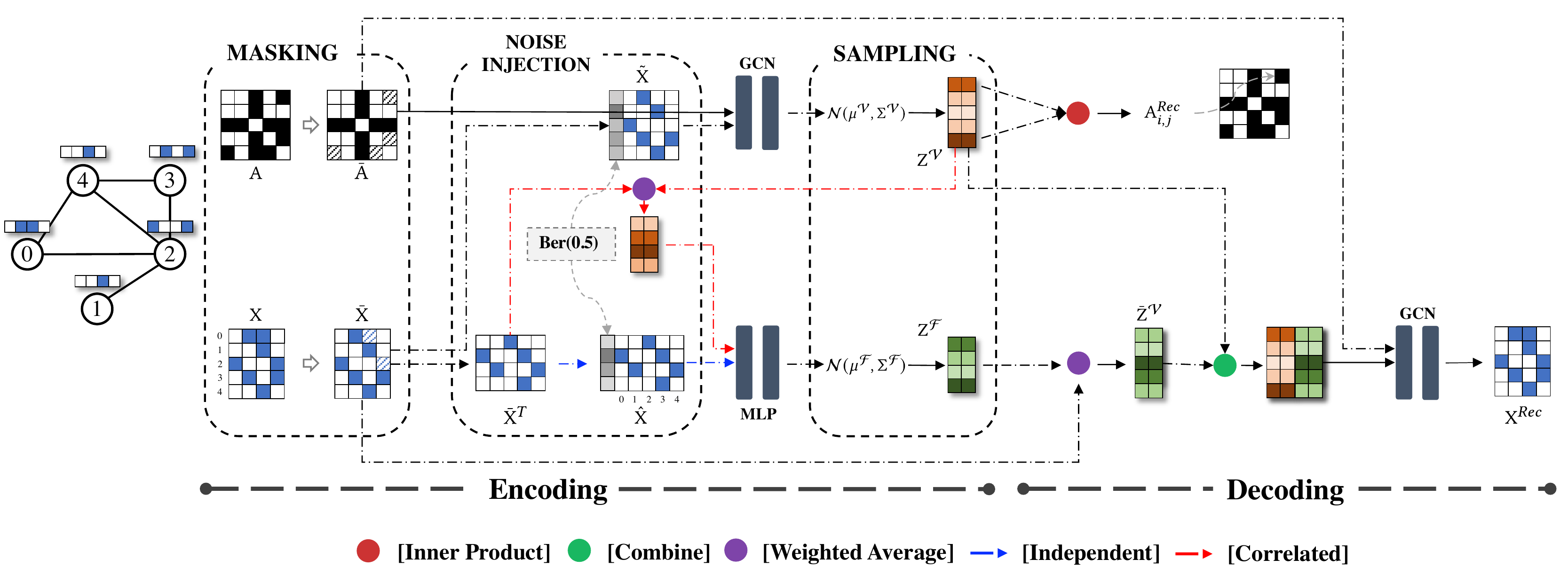}
  \caption{The overall framework of \algo.} 
  \label{fig:framework}
\end{figure*}

In this section,
we present our model \algo. 
Different from SIG-VAE that uses node embeddings only,
\algo\ further generates feature embeddings to capture the rich semantic information on node features,
which can be used to enhance feature reconstruction.
Specifically,
we consider two cases in the encoder when generating node and feature embeddings: (1) they are independent; (2) they are correlated.
After that,
in the decoder part,
we utilize GNNs to reconstruct node features based on both node and feature embeddings.
Finally,
we show how structure/feature masking can be integrated with the hierarchical variational framework
and gives the optimization techniques.
The overall framework of \algo\ is summarized in Figure~\ref{fig:framework}.

\subsection{Variational lower bound}
\label{sec:vlb}
In VI,
given a graph $\mathcal{G}$ with an adjacency matrix $\rma$ and a feature matrix $\rmx$,
we approximate the true posterior $p(\rmzv, \rmza \vert \rma, \rmx)$
with a variational distribution $q(\rmzv, \rmza \vert \psi_1, \psi_2)$,
where $\psi_1$ and $\psi_2$ are variational parameters.
To capture more complex posteriors that go beyond the exponential family,
we adopt the hierarchical variational framework in SIVI and assume
\begin{equation}
\small
\label{eq:hie}
\rmzv \sim q_{1} (\rmzv | \psi_1),\ \psi_1 \sim q_{\phi_1}(\psi_1),\ 
\rmza \sim q_{2} (\rmza | \psi_2),\ \psi_2 \sim q_{\phi_2}(\psi_2),
\end{equation}
where 
$\phi_1$ and $\phi_2$ are parameters of mixing distributions.
We marginalize $\psi_1$ and $\psi_2$ out and derive
\begin{equation}
\label{eq:inte}
\small
\begin{split}
& \rmzv \sim h_{\phi_1}(\rmzv) = \int_{\psi_1} q_{1}(\rmzv |\psi_1)q_{\phi_1}(\psi_1) \rmd \psi_1, \\
& \rmza \sim h_{\phi_2}(\rmza) = \int_{\psi_2} q_{2}(\rmza |\psi_2)q_{\phi_2}(\psi_2) \rmd \psi_2. \\
\end{split}
\end{equation}
We maximize the log-likelihood
of observations $\rma$ and $\rmx$,
and use Jensen's inequality to get
\begin{equation}
\label{eq:elbo}
\log p(\rma, \rmx) \geq \mathbb{E}_{h_{\phi} (\rmzv, \rmza)} \left [ \log \frac{p (\rma,\rmx,\rmzv,\rmza)}{h_{\phi} (\rmzv, \rmza)} \right ] = \mathcal{L},
\end{equation}
where $\mathcal{L}$ is ELBO
and 
\begin{equation}
\label{eq:joint}
h_{\phi}(\rmzv, \rmza) = \int_{\psi_1} \int_{\psi_2} q(\rmzv, \rmza |\psi_1, \psi_2)q_{\phi}(\psi_1, \psi_2) \rmd \psi_1 \rmd \psi_2
\end{equation}
is the marginal distribution over $\rmzv$ and $\rmza$.
Since $h_{\phi}$ is often intractable,
the Monte Carlo estimation of ELBO could be prohibited.
To address the problem,
we first take the mean-field assumption:
\begin{equation}
\label{eq:mean_field}
\begin{split}
& q(\rmzv, \rmza | \psi_1, \psi_2) = q_{1}(\rmzv | \psi_1)  q_{2}(\rmza | \psi_2), \\
& q_{\phi}(\psi_1, \psi_2) = q_{\phi_1}(\psi_1)q_{\phi_2}(\psi_2),
\end{split}
\end{equation}
and substitute Eq.~\ref{eq:mean_field} into Eq.~\ref{eq:joint} to get:
\begin{equation}
\label{eq:ind}
h_{\phi}(\rmzv, \rmza) = h_{\phi_1}(\rmzv)  h_{\phi_2}(\rmza).
\end{equation}
From Eq.~\ref{eq:ind},
we see that $\rmzv$ and $\rmza$ are independent.
Then we 
derive a lower bound for the ELBO based on Eq.~\ref{eq:ind}:
\begin{equation}
\label{eq:underl1}
\begin{split}
 \mathcal{L} & = \mathbb{E}_{h_{\phi_1}(\rmzv)} \mathbb{E}_{h_{\phi_2}(\rmza)} \left [ \log \frac{p (\rma,\rmx,\rmzv,\rmza)}{h_{\phi_1}(\rmzv) h_{\phi_2}(\rmza)} \right ]\\ 
& \geq \mathbb{E}_{\psi_1 \sim q_{\phi_1}(\psi)}\mathbb{E}_{\rmzv \sim q_{1}(\rmzv | \psi_1)}\mathbb{E}_{\psi_2 \sim q_{\phi_2}(\psi)}\mathbb{E}_{\rmza \sim q_{2}(\rmza | \psi_2)}\\
& \quad \quad \left [ \log \frac{p (\rma,\rmx,\rmzv,\rmza)}{q_{1}(\rmzv | \psi_1)q_{2}(\rmza | \psi_2)}\right ] = \underl_1 \\
\end{split}
\end{equation}
Details on the derivation of {Equation~\ref{eq:underl1}} are deferred to Appendix~\ref{app:lb}.
In $\underl_1$,
$q_1$ and $q_2$ are required to be explicit and have analytic density function,
while
$q_{\phi_1}$ and $q_{\phi_2}$ could be implicit but have to be convenient to be sampled from.
Directly optimizing $\underl_1$ by Monte Carlo Estimation is much easier.

However,
in practice,
nodes and their features are highly correlated.
On the one hand, 
node embeddings are generated based on features.
On the other hand,
the semantic information of features are directly reflected by nodes.
Therefore,
the independence between $\rmzv$ and $\rmza$ in Equation~\ref{eq:ind} is inappropriate.
To tackle the issue,
we modify Eq.~\ref{eq:mean_field} into:
\begin{equation}
\label{eq:mean_field1}
\begin{split}
& q(\rmzv, \rmza | \psi_1, \psi_2) = q_{1}(\rmzv | \psi_1)  q_{2}(\rmza | \psi_2), \\
& q_{\phi}(\psi_1, \psi_2) = q_{\phi_2}(\psi_2 \vert \psi_1)q_{\phi_1}(\psi_1),
\end{split}
\end{equation}
which explicitly characterizes the dependence between variational parameters $\psi_1$ and $\psi_2$.
In this way,
$h_{\phi}(\rmzv, \rmza) \neq h_{\phi_1}(\rmzv)  h_{\phi_2}(\rmza)$,
which shows that $\rmzv$ and $\rmza$ are correlated.
Then we can derive another lower bound for the ELBO in Equation~\ref{eq:elbo}:
\begin{equation}
\label{eq:underllcor}
\begin{split}
 \mathcal{L} & \geq \mathbb{E}_{\psi_1 \sim q_{\phi_1}(\psi_1)}\mathbb{E}_{\psi_2 \sim q_{\phi_2}(\psi_2 \vert \psi_1)}
\mathbb{E}_{(\rmzv,\rmza) \sim q_(\rmzv,\rmza | \psi_1,\psi_2)} \\
& \quad \quad \left [ \log \frac{p (\rma,\rmx,\rmzv,\rmza)}{q(\rmzv,\rmza | \psi_1,\psi_2)}\right ] = \underl_2 \\
\end{split}
\end{equation}

\subsection{Encoder}
In the encoder,
we generate $\rmzv$ and $\rmza$ from observations $\rma$ and $\rmx$.
We next show how to generate 
$\rmzv$ and $\rmza$ according to whether they are independent or not.

\textbf{[$\rmzv$ and $\rmza$ are independent].}
To generate $\rmzv$,
we assume that
$q_1(\rmzv \vert \psi_1)
= \prod_{i=1}^n q_1(\rmzv_i \vert \muv_i,\Sigmav_i)$,
where 
$q_1(\rmzv_i \vert \muv_i,\Sigmav_i)\\ $$= \mathcal{N}(\muv_i,\Sigmav_i)$
and 
$\mathcal{N}$ is multivariate Gaussian distribution with mean $\muv_i$ and diagonal co-variance matrix $\Sigmav_i$.
Since $\muv_i$ and $\Sigmav_i$ are random variables,
we draw them by injecting noise $\tilde{\epsilon}$ into a 
GNN model:
\begin{equation}
\tilde{\rmx} = \texttt{CONCAT}(\rmx, \tilde{\epsilon}),\ \tilde{\epsilon} \sim \tilde{q} ({\epsilon}),\ [\muv_i, \Sigmav_i] = \texttt{GNN}_1(\rma, \tilde{\rmx}),
\label{psi_1}
\end{equation}
where 
$\texttt{CONCAT}(\cdot)$ is the concatenation function and $\texttt{GNN}_1(\cdot)$ is a GNN model. 
Note that 
$\tilde{\epsilon}$ is random noise sampled from distribution $\tilde{q}(\epsilon)$, whose row size should be the same as $\rmx$.
The injected noise $\tilde{\epsilon}$ enables
the uncertainty propagation {between neighboring nodes} in the GNN layer,  
which drives the outputs of the GNN to be random variables rather than deterministic values.
Similarly,
for $\rmza$,
we assume 
$q_2(\rmza \vert \psi_2) = \prod_{r=1}^l q_2(\rmza_r \vert \mua_r,\Sigmaa_r)$
with 
$q_2(\rmza_r \vert \mua_r,\Sigmaa_r) = \mathcal{N}(\mua_r,\Sigmaa_r)$.
To infer $\mua_r$ and $\Sigmaa_r$,
we use a MLP model:
\begin{equation}
\hat{\rmx} = \texttt{CONCAT}(\rmx^{T}, \hat{\epsilon}),\ \hat{\epsilon} \sim \hat{q} ({\epsilon}),\ [\mua_r, \Sigmaa_r] = \texttt{MLP}_1(\hat{\rmx}).
\label{psi_2}
\end{equation}
Note that $\rmx^T \in \mathbb{R}^{l\times n}$
and the $r$-th row $\rmx^T_r \in \mathbb{R}^n$ can be considered as the feature vector of feature $f_r$.
The random noise $\hat{\epsilon}$ drawn from $\hat{q}(\epsilon)$ injects uncertainty to the matrix $\rmx^{T}$,
which models $\mua_r$ and $\Sigmaa_r$ as random variables.

\textbf{[$\rmzv$ and $\rmza$ are correlated].}
We also assume $q_1$ and $q_2$ follow Gaussian distribution
and use the same method as in Equation~\ref{psi_1} to generate $[\muv_i, \Sigmav_i]$.
However,
to capture the dependence between $\psi_1$ and $\psi_2$~\footnote{Here,
we denote $\psi_1 = [\muv, \Sigmav]$ and $\psi_2 = [\mua, \Sigmaa]$, respectively.} in Equation~\ref{eq:mean_field1},
we compute $[\mua_r, \Sigmaa_r]$
for feature $f_r$ based on node embeddings.
Specifically,
since the rich semantic information of each feature is directly reflected by values of nodes in the feature,
we take the feature vector 
$\rmx^T_r \in \mathbb{R}^n$ as the weight vector over all the nodes,
and compute:
\begin{equation}
\label{eq:cor}
[\mua_r, \Sigmaa_r] = \texttt{MLP}_2\left(\frac{\sum_{i=1}^n \rmx^T_{ri} [\muv_i, \Sigmav_i]}{\sum_{i=1}^n \rmx^T_{ri}}\right ).  
\end{equation}
In this way,
$[\mua_r, \Sigmaa_r]$ is derived from node embeddings.
Since $\muv_i$ and $\Sigmav_i$ are random variables,
$\mua_r$ and $\Sigmaa_r$ will also be random variables.

\subsection{Decoder}
In the decoder,
we aim to reconstruct both edges and features in the given graph.
The generative process is described as follows.

First, for each node $x_i$ and each feature $f_r$, 
we draw $(\rmzv_i,\rmza_r) \sim h_{\phi}(\rmzv_i,\rmza_r)$\footnote{
When $\rmzv_i$ and $\rmza_r$
are independent, we draw
$\rmzv_i \sim h_{\phi_1}(\rmzv_i)$ and $\rmza_r \sim h_{\phi_2}(\rmza_r)$.}.
Second,
for each edge $\rma_{ij}$ in the adjacency matrix $\rma$, 
draw $\rma_{ij} \sim \texttt{Ber}(p_{ij}^{\rma})$.
Here,
$\texttt{Ber}(\cdot)$ denotes Bernoulli distribution and 
$p_{ij}^{\rma}$ is the probability for the existence of edge $\rma_{ij}$.
We implement $p_{ij}^{\rma}$ 
simply by inner product as:
$p_{ij}^{\rma} = \sigma ((\rmzv_i)^T \rmzv_j)$,
where $\sigma$ is the sigmoid function.
Third,
for each attribute $\rmx_{ir}$ in the attribute matrix $\rmx$, draw $\rmx_{ir} \sim \mathcal{N}(\mu_{ir}^{\rmx}, \Sigma_{ir}^{\rmx} \vert \rmzv_i, \rmza_r)$.
Here, 
$\mu_{ir}^{\rmx}, \Sigma_{ir}^{\rmx}$ are functions of $\rmzv$ and $\rmza$.

We next introduce how to compute 
$\mu_{ir}^{\rmx}$ and $\Sigma_{ir}^{\rmx}$.
Since $\rmza_r$ contains rich semantic information on feature $f_r$,
the affinity between $x_i$ and $f_r$
can provide fine-grained knowledge for feature reconstruction.
Given a node $x_i$,
to capture its affinities with all the features,
the attention mechanism~\cite{vaswani2017attention}
can be applied on node and feature embeddings. 
However,
this will increase the time complexity of the model.
For simplicity,
we directly use
the feature vector $\rmx_i \in \mathbb{R}^l$ of $x_i$
as the weight vector and calculate the weighted average over all the feature embeddings:
\begin{equation}
\bar{\mathrm{Z}}^{\mathcal{V}}_i = \frac{\sum_{r=1}^l \rmx_{ir}\rmza_r}{\sum_{r=1}^l \rmx_{ir}}.
\end{equation}
Compared with $\rmzv_i$,
$\bar{\mathrm{Z}}^{\mathcal{V}}_i$
contains more details on how each feature can be reconstructed.
After that,
we combine $\rmzv_i$ and
$\bar{\mathrm{Z}}^{\mathcal{V}}_i$
to get:
\begin{equation}
\rmzv_i = \texttt{COMBINE}(\rmzv_i, \bar{\mathrm{Z}}^{\mathcal{V}}_i).
\end{equation}
In our experiments,
we set the \texttt{COMBINE} function to be \texttt{CONCAT}.
Finally, 
the updated $\rmzv_i$ is taken as input and fed into a GNN model to learn parameters w.r.t. node $x_i$:
\begin{equation}
[\mu_{i}^{\rmx}, \Sigma_{i}^{\rmx}] = \texttt{GNN}_2(\rma,\rmzv_i),
\end{equation}
where $\texttt{GNN}_2(\cdot)$ is a GNN model.

\subsection{Masking}
To further improve the model generalizability,
we introduce the masking mechanism 
in \algo\ by adding an additional layer to the hierarchical variational framework.
Specifically,
we transform Equation~\ref{eq:joint} into:
\begin{equation}
\label{eq:joint1}
\begin{split}
h_{\phi}(\rmzv, \rmza) & = \int_{\psi_1} \int_{\psi_2} \int_{\tilde{G}} q(\rmzv, \rmza |\psi_1, \psi_2) \cdot \\
& \quad \quad \quad \quad \quad q_{\phi}(\psi_1, \psi_2|\tilde{G})p(\tilde{G}|\rma,\rmx) \rmd \psi_1 \rmd \psi_2 \rmd \tilde{G}\\
\end{split}
\end{equation}
From the above equation,
we see that 
in addition to $\psi_1$ and $\psi_2$,
the integration is performed over 
a new variable $\tilde{G}$ 
and a probability function $p(\tilde{G}|\rma,\rmx)$.
Here,
$\tilde{G}$ denotes a new graph and $p$ is the graph augmentation probability function.
The equation
can lead to a new 
variational lower bound 
for the ELBO,
but it is more difficult to optimize compared with $\underl_1$ and $\underl_2$.
To tackle the issue,
we  can first perform graph augmentation and generate a perturbed graph $\tilde{G}$.
After that,
based on $\tilde{G}$,
node and feature embeddings are learned based on $\underl_1$ or $\underl_2$.
We repeat the above process 
until convergence.
Although graph augmentation can include more operations than  masking,
we mainly focus on structure/feature masking in this paper,
because
masking is beneficial for node classification~\cite{hou2022graphmae}.

\subsection{Optimization}
In Section~\ref{sec:vlb},
we have derived two lower bounds $\underl_1$ and $\underl_2$ for the ELBO, according to whether $\rmzv$ and $\rmza$ are independent.
For notation brevity,
we use $\underl$ to overload both $\underl_1$ and $\underl_2$.
However,
directly optimizing $\underl$ 
could lead to the degeneracy problem~\cite{pmlr-v80-yin18b} that
$q_{\phi_1}(\psi_1)$,
$q_{\phi_2}(\psi_2)$
and $q_{\phi}(\psi_1,\psi_2)$
might converge to a point mass density, 
which degenerates SIVI to the vanilla VI.
To address the problem, 
we can regularize $\underl$ by $B_K$:
{\small{
\begin{equation}
\nonumber
B_K  = \mathbb{E}_{(\psi_1, \psi_2 ), \{(\tilde{\psi}_1^{k}, \tilde{\psi}_2^{k})\}_{k=1}^K \sim q_{\phi}(\psi_1, \psi_2)}  \mathrm{D_\text{KL}}\left (q(\rmzv,\rmza \vert \psi_1, \psi_2) || \tilde{h}_K(\rmzv, \rmza)\right )
\end{equation}
}}
\begin{equation}
\nonumber
\tilde{h}_K(\rmzv, \rmza) = \frac{q(\rmzv,\rmza | \psi_1, \psi_2) + \sum_{k=1}^K q(\rmzv, \rmza | \tilde{\psi}_1^{k}, \tilde{\psi}_2^{k}) }{K+1}.
\end{equation}
Note that $B_K$ satisfies (1) $B_K \geq 0$; (2) $B_K = 0$ if and only if $K = 0$ or $q_{\phi}$ degenerates to a point mass density.
According to~\cite{pmlr-v80-yin18b},
$\underl_K = \underl + B_K$
is an asymptotically exact surrogate ELBO that satisfies $\underl_0 = \underl$ and $\lim_{K\rightarrow \infty}\underl_K = \mathcal{L}$.
Maximizing $\underl_K$ with $K\geq 1$ derives positive $B_K$ and could drive $q_{\phi}$ away from degeneracy.
Moreover,
importance reweighting~\cite{burda2015importance} can be further introduced to tighten $\underl_K$
by drawing $J$ samples $\{(\rmzv)_j, (\rmza)_j, \psi_1^j, \psi_2^j\}_{j=1}^J$ from $q(\rmzv, \rmza , \psi_1, \psi_2)$.
The objective can be formulated as
\begin{equation}
\small
\label{eq:reweight}
\begin{split}
\underl_K^{J} 
= & \mathbb{E}_{\{(\rmzv)_j, (\rmza)_j, \psi_1^j, \psi_2^j\}_{j=1}^J \sim q(\rmzv, \rmza | \psi_1, \psi_2) q_{\phi}(\psi_1, \psi_2) } \\
&   \mathbb{E}_{\{\tilde{\psi}_1^{k}, \tilde{\psi}_2^{k}\}_{k=1}^K \sim q_{\phi}(\psi_1, \psi_2)} \log \frac{1}{J} \sum_{j=1}^J \frac{p(\rma, \rmx, (\rmzv)_j, (\rmza)_j)}{\Omega_j},\\
\end{split}
\end{equation}
where
{\small{
\begin{equation}
\nonumber
\Omega_j = \frac{1}{K+1}\left[ q((\rmzv)_j, (\rmza)_j | \psi_1^j, \psi_2^j) + \sum_{k=1}^K q((\rmzv)_j, (\rmza)_j | \tilde{\psi}_1^{k}, \tilde{\psi}_2^{k}) \right].
\end{equation}
}}
Then
we take
$\underl_K^{J} $ as the surrogate ELBO and use stochastic gradient ascent to optimize it.
Note that $\underl_1$ and $\underl_2$ treats differently for $q_{\phi}(\psi_1,\psi_2)$.
Finally,
we 
summarize the pseudocodes of \algo\ in Algorithm~\ref{alg:hoane} (see Appendix~\ref{app:pseudo}).

[\textbf{Complexity analysis}]
The major time complexity in the encoder comes from GNN and MLP.
Suppose we use GCN as the GNN model.
Since the adjacency matrix is generally sparse,
let $d_{A}$ be the average number of non-zero entries in each row of the adjacency matrix.
Let $l$ be the number of features
and $d$ be the embedding dimension.
Further,
we denote $\tilde{d}$ and $\hat{d}$ as the dimensions of injected noise to the GCN and MLP, respectively.
Then,
the time complexities for GCN and MLP are
$O (nd_A(l+\tilde{d}) + n (l+\tilde{d}) d )$
and
$O(l(n+\hat{d})d)$,
respectively.
In the decoder,
suppose we still adopt GCN as the GNN model.
Then the time complexities for reconstructing links and features are $O(n^2d)$ and $O (nd_Ad + ndl )$, respectively.
As suggested by~\cite{kipf2016variational},
we can down-sample the number of nonexistent edges in the graph to reduce the time complexity for recovering links.




%% file: tex/experiment.tex
\section{Experiment}
\label{sec:exp}
In this section
we comprehensively evaluate the quality of node embeddings learned by \algo.
We mainly study four research questions:

\textbf{(RQ1)} How does \algo\ perform in the link prediction task?

\textbf{(RQ2)} Can \algo\ effectively predict node attributes?

\textbf{(RQ3)} While \algo\ is an unsupervised learning method,
can it perform well when generalized to the node classification task?

\textbf{(RQ4)} How does structure/feature masking influence the performance of \algo?


\subsection{Datasets and Baselines}
To answer the above four questions,
we conduct extensive experiments on seven public datasets:
\emph{Cora}, \emph{Citeseer},
\emph{Pubmed},
\emph{Coauthor CS}, \emph{Coauthor Physics}, \emph{Amazon Computer} and \emph{Amazon Photo}.
Detailed descriptions and statistics on these datasets are provided in Appendix~\ref{app:data}.
We also compare \algo\ with 9 other SOTA baselines, 
which can be categorized into two groups:

\noindent{\small$\bullet$}\textbf{[Generative graph SSL methods]}.
This group of methods are based on GAE/VGAE and aim to reconstruct links and/or features,
including
SIG-VAE~\cite{hasanzadeh2019semi}, 
CAN~\cite{meng2019co}, GATE~\cite{salehi2019graph}
and GraphMAE~\cite{hou2022graphmae}.
Note that GraphMAE is the SOTA generative graph SSL model.

\noindent{\small$\bullet$}\textbf{[Graph contrastive learning methods]}.
Models in this type construct positive (and negative) pairs for contrast to learn node representations,
including DGI~\cite{velickovic2019deep}, MVGRL~\cite{hassani2020contrastive}, 
GRACE~\cite{zhu2020deep}, GCA~\cite{zhu2021graph} and CCA-SSG~\cite{zhang2021canonical}.

Further,
for more implementation details, 
see Appendix~\ref{app:imp}.

\begin{table*}[!htbp]
\centering
\caption{Link prediction results. The error bar $(\pm)$ denotes the standard deviation score of results over 10 trials. 
We highlight the best score on each dataset in bold.
For CAN, 
the released codes by the authors do not implement reconstruction for numerical features, 
so we cannot run it on datasets with numerical features.
OOM denotes the out-of-the-memory error.
}
\resizebox{0.95\linewidth}{!}
{
\begin{tabular}{c|c|ccccccc} \hline
Metrics & Method & Cora & Citeseer & Pubmed & Photo & Computer & CS & Physics \\ \hline 
\multirow{9}{*}{AUC} 
& DGI       & $93.88 \pm 1.00$ & $95.98 \pm 0.72$ & $96.30 \pm 0.20$ & $80.95 \pm 0.39$ & $81.27 \pm 0.51$ & $93.81 \pm 0.20$ & $93.51 \pm 0.22$ \\
& MVGRL     & $93.33 \pm 0.68$ & $88.66 \pm 5.27$ & $95.89 \pm 0.22$ & $69.58 \pm 2.04$ & $92.37 \pm 0.78$ & $91.45 \pm 0.67$ & OOM \\
& GRACE     & $82.67 \pm 0.27$ & $87.74 \pm 0.96$ & $94.09 \pm 0.92$ & $81.72 \pm 0.31$	& $82.94 \pm 0.20$ & $85.26 \pm 2.07$ & $83.48 \pm 0.96$\\
& GCA       & $81.46 \pm 4.86$ & $84.81 \pm 1.25$ & $94.20 \pm 0.59$ & $70.02 \pm 9.66$ & $89.92 \pm 0.91$ & $84.35 \pm 1.13$ & $85.24 \pm 5.41$\\
& CCA-SSG   & $93.88 \pm 0.95$ & $94.69 \pm 0.95$ & $96.63 \pm 0.15$ & $73.98 \pm 1.31$ & $75.91 \pm 1.50$ & $96.80 \pm 0.16$ & $96.74 \pm 0.05$\\
& CAN       & $93.67 \pm 0.62$ & $94.56 \pm 0.68$ & $-$ & $97.00 \pm 0.28$ & $96.03 \pm 0.37$ & $-$ & $-$ \\ 
& SIG-VAE   & $94.10 \pm 0.68$ & $92.88 \pm 0.74$ & $85.89 \pm 0.54$ & $94.98 \pm 0.86$ & $91.14 \pm 1.10$ & $95.26 \pm 0.36$ & $98.76 \pm 0.23$ \\
& GraphMAE  & $90.70 \pm 0.01$ & $70.55 \pm 0.05$ & $69.12 \pm 0.01$ & $77.42 \pm 0.02$ & $75.14 \pm 0.02$ & $91.47 \pm 0.01$ & $87.61 \pm 0.02$\\ 
& \algo-v1  & $94.95 \pm 0.72$  & $96.75 \pm 0.54$  & $97.07 \pm 2.20$  & $98.40 \pm 0.08$ & $96.87 \pm 0.29$ & $97.82 \pm 0.11$ & $98.95 \pm 0.06$ \\
& \algo-v2  & $95.37 \pm 0.60$ & $96.81 \pm 0.51$ & $97.79 \pm 0.22$ & $98.47 \pm 0.05$ & $97.28 \pm 0.00$ & $97.83 \pm 0.11$ & $98.97 \pm 0.04$ \\
& \algo-v3  & $\bm{95.50 \pm 0.71}$ & $\bm{97.04 \pm 0.47}$ & $\bm{97.87 \pm 0.20}$ & $\bm{98.64 \pm 0.05}$ & $\bm{97.70 \pm 0.19}$ & $\bm{98.42 \pm 0.13}$ & $\bm{99.03 \pm 0.05}$ \\
\hline
\multirow{9}{*}{AP} 
& DGI       & $93.60 \pm 1.14$ & $96.18 \pm 0.68$ & $95.65 \pm 0.26$ & $81.01 \pm 0.47$ & $82.05 \pm 0.50$ & $92.79 \pm 0.31$ & $92.10 \pm 0.29$\\
& MVGRL     & $92.95 \pm 0.82$ & $89.37 \pm 4.55$ & $95.53 \pm 0.30$ & $63.43 \pm 2.02$ & $91.73 \pm 0.40$ & $89.14 \pm 0.93$ & OOM \\
& GRACE     & $82.36 \pm 0.24$ & $86.92 \pm 1.11$ & $93.26 \pm 1.20$ & $81.18 \pm 0.37$	& $83.12 \pm 0.23$ & $83.90 \pm 2.20$ & $82.20 \pm 1.06$ \\
& GCA       & $80.87 \pm 4.11$ & $81.93 \pm 1.76$ & $93.31 \pm 0.75$ & $65.17 \pm 10.11$ & $89.50 \pm 0.64$ & $83.24 \pm 1.16$ & $82.80 \pm 4.46$\\
& CCA-SSG   & $93.74 \pm 1.15$ & $95.06 \pm 0.91$ & $95.97 \pm 0.23$ & $67.99 \pm 1.60$ & $69.47 \pm 1.94$ & $96.40 \pm 0.30$ & $96.26 \pm 0.10$\\
& CAN       & $94.49 \pm 0.60$ & $95.49 \pm 0.61$ & $-$ & $96.68 \pm 0.30$ & $95.96 \pm 0.38$ & $-$ & $-$ \\ 
& SIG-VAE   & $94.79 \pm 0.71$ & $94.21 \pm 0.53$ & $85.02 \pm 0.49$ & $94.53 \pm 0.93$ & $91.23 \pm 1.04$ & $94.93 \pm 0.37$ & $98.85 \pm 0.12$ \\
& GraphMAE  & $89.52 \pm 0.01$ & $74.50 \pm 0.04$ & $87.92 \pm 0.01$ & $77.18 \pm 0.02$	& $75.80 \pm 0.01$ & $83.58 \pm 0.01$ & $86.44 \pm 0.03$\\
& \algo-v1  & $95.53 \pm 0.54$ & $97.10 \pm 0.49$ & $97.25 \pm 2.07$ & $98.32 \pm 0.09$ & $96.73 \pm 0.31$ & $98.30 \pm 0.11$ & $99.10 \pm 0.09$ \\ 
& \algo-v2  & $95.90 \pm 0.49$ & $97.17 \pm 0.46$ & $\bm{97.89 \pm 0.21}$ & $98.37 \pm 0.09$ & $97.15 \pm 0.00$ & $98.33 \pm 0.10$ & $99.13 \pm 0.06$ \\
& \algo-v3  & $\bm{95.92 \pm 0.68}$ & $\bm{97.33 \pm 0.46}$ & $97.87 \pm 0.20$ & $\bm{98.48 \pm 0.06}$ & $\bm{97.50 \pm 0.15}$ & $\bm{98.53 \pm 0.18}$ & $\bm{99.18 \pm 0.04}$ \\
\hline
\end{tabular}
}
\label{results:lp}
\end{table*}
\begin{table*}[h]
\centering
\caption{Attribute inference performance w.r.t the MSE metric.
The best result in each dataset is highlighted in bold.
}
\resizebox{\linewidth}{!}
{
\begin{tabular}{c|ccccccc} \hline
Method & Cora & Citeseer & Pubmed & Photo & Computer & CS & Physics \\ 
\hline 
CAN & - & - & - & $0.22 \pm 0.00$ & $0.23 \pm 0.01$ & -& - \\
GATE    & $1.80 \times 10^{-3} \pm 2.15 \times 10^{-4}$ & {$\bm{4.58 \times 10^{-4} \pm 8.07 \times 10^{-5}}$} & $3.90 \times 10^{-4} \pm 1.99 \times 10^{-5}$ & $0.24 \pm 0.01$ & $0.25 \pm 0.01$ & $2.03 \pm 0.25$ &  OOM \\
GraphMAE& {$\bm{1.57 \times 10^{-3} \pm 7.42 \times 10^{-5}}$} & $8.68 \times 10^{-4} \pm 1.30 \times 10^{-4}$ & $7.29 \times 10^{-4} \pm 2.66 \times 10^{-5}$ & $0.48 \pm 0.00$ & $0.48 \pm 0.00$ & $2.70 \pm 0.06$ & $2.97 \pm 0.05$\\
\algo-v1& $1.90 \times 10^{-3} \pm 8.18 \times 10^{-5}$ & $4.87 \times 10^{-4} \pm 2.62 \times 10^{-6}$ & $4.21 \times 10^{-4} \pm 4.70 \times 10^{-5}$ & $0.22 \pm 0.00$ & $0.23 \pm 0.14$ & $2.12 \pm 0.06
$ & $2.15 \pm 0.05$\\ 
\algo-v2& $1.89 \times 10^{-3} \pm 8.13 \times 10^{-5}$ & $4.86 \times 10^{-4} \pm 2.24 \times 10^{-6}$ & $3.68 \times 10^{-4} \pm 7.11 \times 10^{-6}$ & $\bm{0.21 \pm 0.01}$ & $0.23 \pm 0.01$ & $2.08 \pm 0.07$ & $\bm{2.14 \pm 0.03}$\\
\algo-v3& $1.89 \times 10^{-3} \pm 8.13 \times 10^{-5}$ & $4.84 \times 10^{-4} \pm 2.96 \times 10^{-6}$ & $\bm{3.66 \times 10^{-4} \pm 7.34 \times 10^{-6}}$ & $\bm{0.21 \pm 0.01}$ & $\bm{0.22 \pm 0.00}$ & $\bm{1.93 \pm 0.07}$ & $\bm{2.14 \pm 0.03}$\\
\hline
\end{tabular}
}
\label{results:ai}
\end{table*}

\begin{table*}[h]
\centering
\caption{Node classification performance w.r.t. the classification accuracy. 
We highlight the best results in bold.
}
\resizebox{0.85\linewidth}{!}
{
\begin{tabular}{c|ccccccc} \hline
Method & Cora & Citeseer & Pubmed & Photo & Computer & CS & Physics \\ 
\hline
DGI         & $82.3 \pm 0.6$    & $71.8 \pm 0.7$    & $76.8 \pm 0.6$    & $91.61 \pm 0.22$ & $83.95 \pm 0.47$ & $92.15 \pm 0.63$ & $94.51 \pm 0.52$ \\ 
MVGRL       & $83.5 \pm 0.4$    & $73.3 \pm 0.5$    & $80.1 \pm 0.7$    & $91.74 \pm 0.07$ & $87.52 \pm 0.11$ & $92.11 \pm 0.12$ & $95.33 \pm 0.03$ \\ 
GRACE       & $81.9 \pm 0.4$    & $71.2 \pm 0.5$    & $80.6 \pm 0.4$    & $92.15 \pm 0.24$ & $86.25 \pm 0.25$ & $92.93 \pm 0.01$ & $95.26 \pm 0.02$ \\ 
CCA-SSG     & $84.0 \pm 0.4$    & $73.1 \pm 0.3$    & $81.0 \pm 0.4$    & \bm{$93.14 \pm 0.14$} & \bm{$88.74 \pm 0.28$} & $93.31 \pm 0.22$ & $95.38 \pm 0.06$ \\ 
GraphMAE    & $84.2 \pm 0.4$    & $\bm{73.4 \pm 0.4}$    & $\bm{81.1 \pm 0.4}$    & $92.98 \pm 0.35$ & $88.34 \pm 0.27$ & $93.08 \pm 0.17$ & $95.30 \pm 0.12$\\ 
\algo-v1   & $82.9 \pm 0.4$  & $71.7 \pm 0.6$  & $78.9 \pm 0.9$  & $92.53 \pm 0.41$ & $88.44 \pm 0.24$ & $93.72 \pm 0.29$ & $\bm{95.40 \pm 0.10}$\\ 
\algo-v2   & $84.0 \pm 0.4$  & $73.0 \pm 0.8$  & $80.4 \pm 0.4$  & $92.70 \pm 0.42$ & $88.39 \pm 0.26$ & $93.83 \pm 0.22$ & $95.39 \pm 0.08$\\ 
\algo-v3   & $\bm{84.3 \pm 0.4}$  & $73.0 \pm 0.8$  & $80.4 \pm 0.4$  & $92.81 \pm 0.45$ & $88.39 \pm 0.26$ & \bm{$93.84 \pm 0.11$} & $95.39 \pm 0.08$\\ 
\hline
\end{tabular}
}
\label{results:cla}
\end{table*}

\begin{figure*}[h]
\centering
\subfigure[Cora-AUC]{
\includegraphics[width=0.22\linewidth]{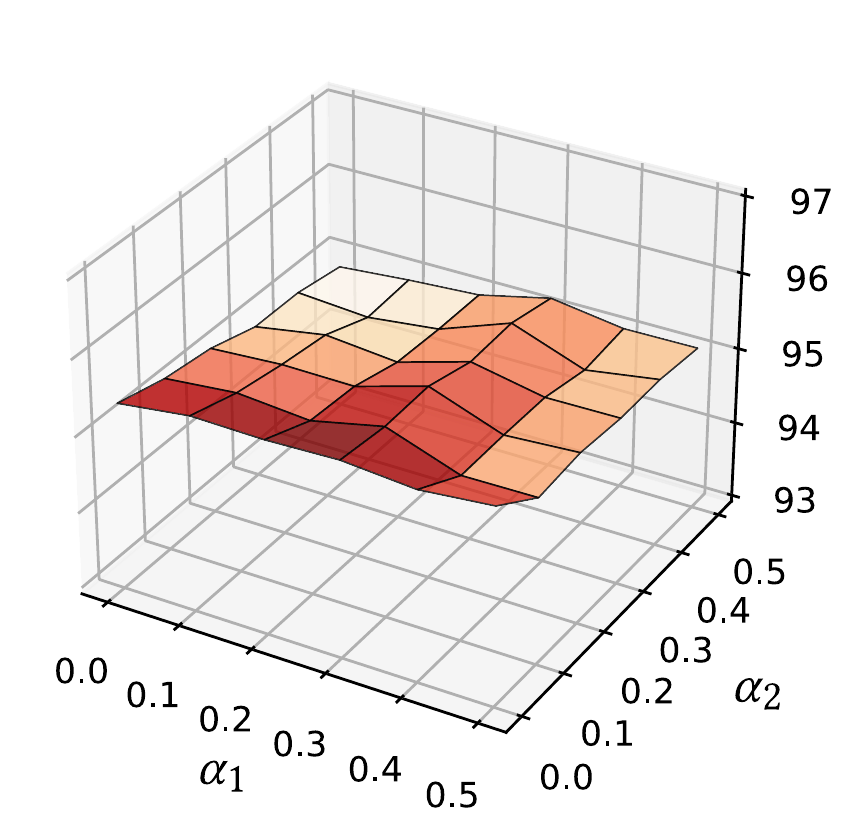}}
\subfigure[Cora-AP]{
\includegraphics[width=0.22\linewidth]{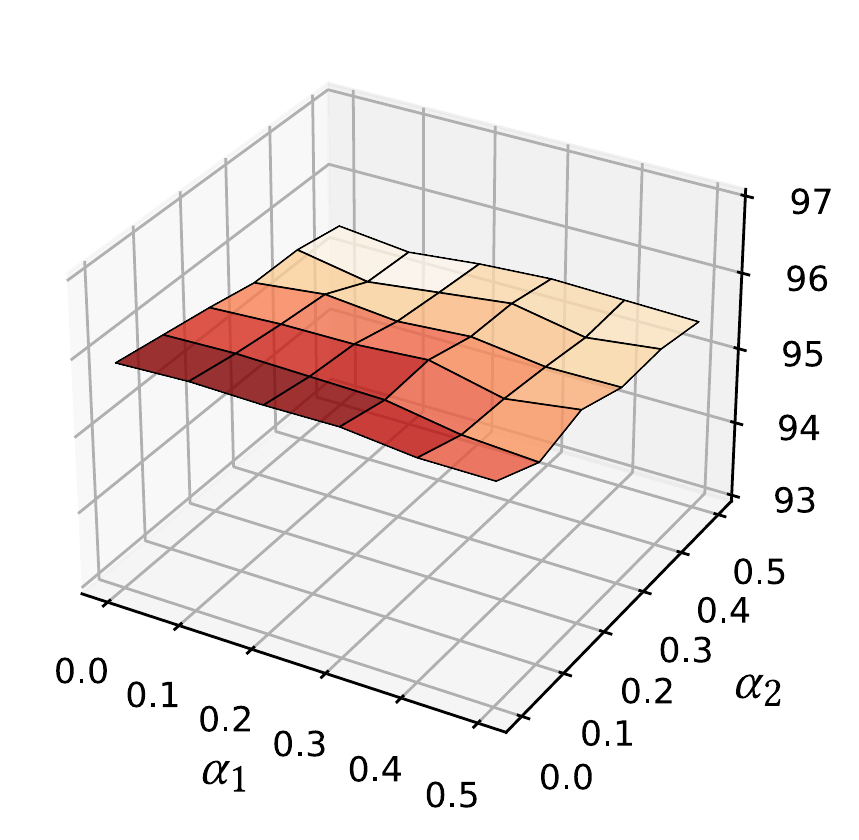}}
\subfigure[Citeseer-AUC]{
\includegraphics[width=0.22\linewidth]{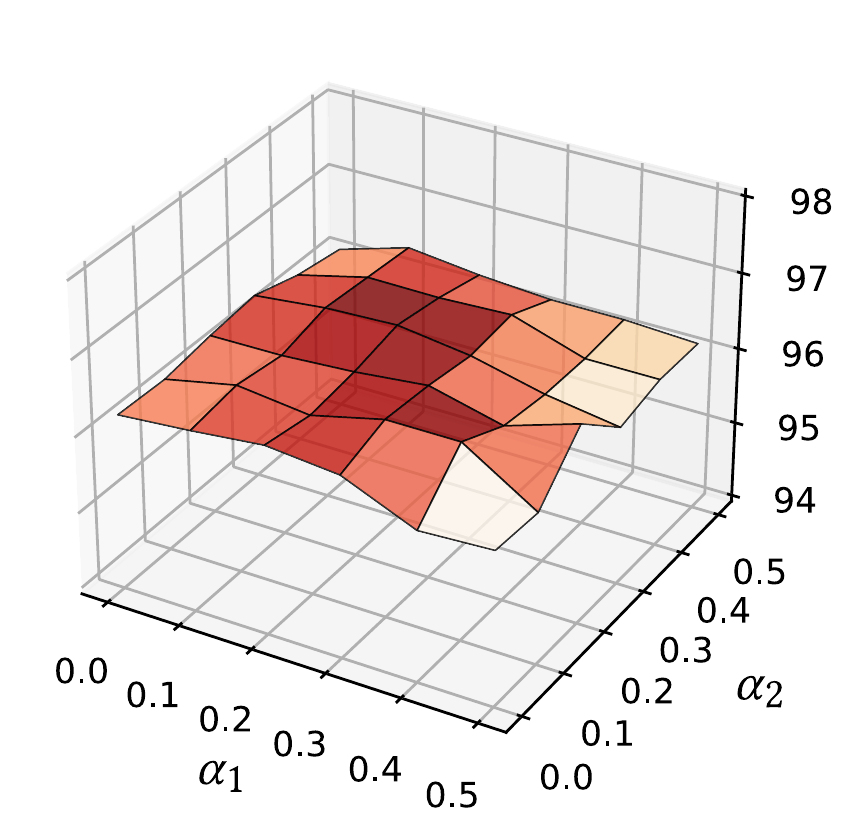}}
\subfigure[Citeseer-AP]{
\includegraphics[width=0.22\linewidth]{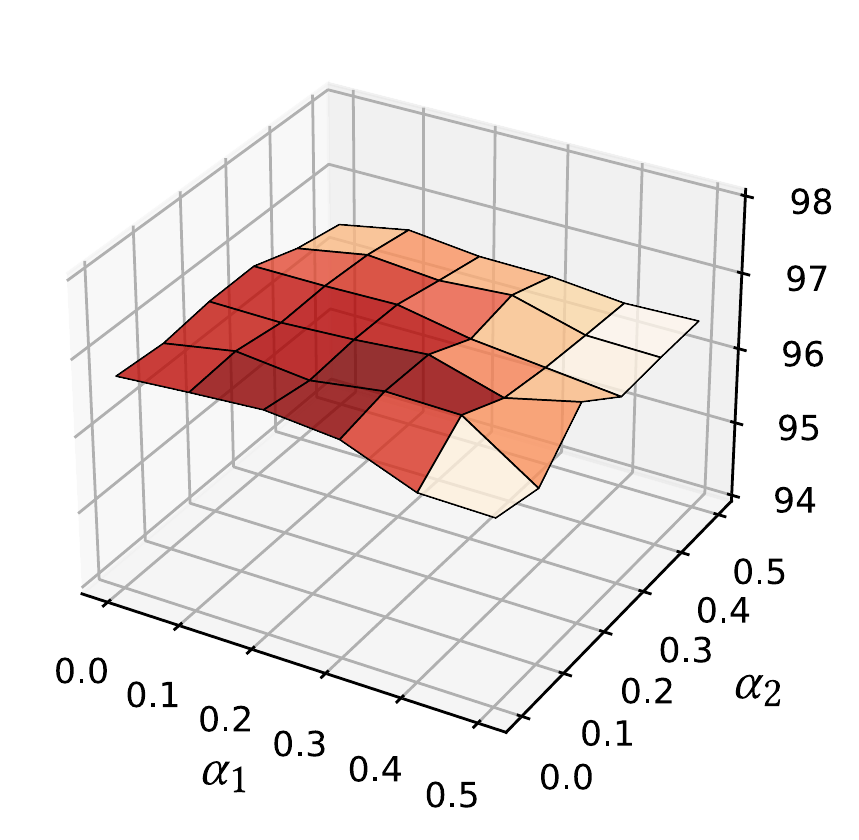}}
\caption{Hyper-parameter sensitivity analysis on the masking rates $\alpha_1$ and $\alpha_2$ in terms of link prediction. The darker the color, the larger the value.}
\label{fig:sensi}
\end{figure*}

\subsection{Link Prediction (RQ1)}
Link prediction is a typical unsupervised learning task for graph analysis,
which aims to predict whether an edge exists between two nodes or not.
We compare \algo\ with 8 other SOTA baselines, 
including GCL models: DGI~\cite{velickovic2019deep}, MVGRL~\cite{hassani2020contrastive}, GRACE~\cite{zhu2020deep}, GCA~\cite{zhu2021graph}, 
CCA-SSG~\cite{zhang2021canonical},
and the generative graph SSL methods: CAN~\cite{meng2019co}, SIG-VAE~\cite{hasanzadeh2019semi}, GraphMAE~\cite{hou2022graphmae}.
For our proposed method \algo,
we put forward three versions.
Specifically,
\algo-v1 assumes the independence between $\rmzv$ and $\rmza$ and optimizes $\underl_1$,
while \algo-v2 captures the correlations between them and optimizes $\underl_2$.
Further,
\algo-v3 upgrades \algo-v2 by adding the masking mechanism.

To evaluate the model performance,
we construct the validation/test set by randomly selecting $20\%/10\%$ edges in the original graph as positive samples
and an equal number of nonexistent edges as negative samples.
After the removal of these selected edges,
we train all the models on the resulting graph with the remaining $70\%$ edges.
We use two commonly used metrics, 
the area under the ROC curve (AUC) and the average precision (AP), to report the model performance.
For both metrics,
a larger value indicates a better performance.
We use the validation set for hyper-parameter tuning and early stopping with a patience of 100,
i.e.,
we stop training if both metric scores on the validation set do not increase for 100 consecutive epochs.
Similar as in~\cite{kipf2016variational},
the predicted probability of an edge between nodes $x_i$ and $x_j$ is calculated by $\rma_{ij} \sim \texttt{Ber}(p_{ij}^{\rma})$,
where $p_{ij}^{\rma} = \sigma ((\rmzv_i)^T \rmzv_j)$
and $\sigma$ is the sigmoid function.
For each method, we run experiments 10 times and report the average results on the test set.
Table~\ref{results:lp}
summarizes the results across all the datasets.
From the table, we have the following observations:

(1)
The generative graph SSL methods except GraphMAE generally perform better than GCL methods.
This is because these methods learn to reconstruct links in the objective. 
For GraphMAE,
it only reconstructs features, which explains its poor performance.

(2)
While \algo\ is based on SIG-VAE,
it achieves better performance. 
This demonstrates the importance of 
feature reconstruction and structure/feature masking. 

(3)
Although CAN co-embeds both nodes and features,
it still performs not very well,
due to 
the independence assumption between node and feature embeddings.
Further,
it uses linear decoder for feature reconstruction,
which restricts the model's effectiveness.

(4)
\algo\ significantly outperforms other competitors across all the datasets,
which indicates the superiority of generative VGAE model in graph representation learning.
In particular,
the consistent outperformance of \algo-v2 over \algo-v1 verifies the importance of capturing the correlations between node and feature embeddings.
Further,
the improvement of \algo-v3 over \algo-v2 shows the necessity of structure/feature masking.

\subsection{Attribute Inference (RQ2)}
Attribute inference is a task that predicts values of missing node attributes.
Similar as in link prediction,
we hide a certain percentage of node features and train on the rest.
To construct the training/validation/test set,
we randomly select $70\%/10\%/20\%$ node features.
The validation set is used for hyper-parameter tuning and early stopping with a patience of 100 epoches.
We take the Mean Squared Error (MSE) as the evaluation metric.
The smaller the value,
the better the performance.
In this task,
we compare \algo\ with 
generative graph SSL methods that reconstruct features in their decoders,
including
CAN~\cite{park2019symmetric},
GATE~\cite{salehi2019graph}
and GraphMAE~\cite{hou2022graphmae}.
For GCL models and other generative graph SSL methods that recover links only,
they cannot be easily adapted to the task, so we do not take them as baselines.
For each method, we run experiments 10 times and report the average results in Table~\ref{results:ai}.
For Cora and Citeseer, we normalize node features for fair comparison, so CAN cannot be applied.
From the table,
while CAN, GATE and GraphMAE can perform well on some datasets,
they cannot consistently provide excellent performance.
For example,
GraphMAE achieves the best result on Cora, but it performs very poorly on Citeseer.
Further,
\algo-v3 outperforms other competitors on 5 out of 7 datasets.
This shows the effectiveness of our proposed feature reconstruction method and also the masking mechanism.
We also notice that,
in all cases, 
\algo-v2 achieves better performance than \algo-v1,
which again verifies the necessity of capturing correlations between node and feature embeddings.


\subsection{Node Classification (RQ3)}
To further study \algo,
we generalize learned embeddings to the node classification task.
After node embeddings are trained on the entire graph,
we train an additional classifier.
Here, we employ Logistic Regression as the classifier.
For Cora, Citeseer and Pubmed, 
we use the public split for evaluation, where each class has fixed 20 nodes for training, another
fixed 500 nodes and 1000 nodes for validation and testing, respectively.
For other datasets,
we randomly split the nodes into 10\%/10\%/80\% training/validation/test sets.
We use classification accuracy as the metric
to evaluate the model performance.
Since GCL methods have been shown to perform well in classification tasks,
we compare \algo\ with 4 state-of-the-arts, 
including DGI, MVGRL, GRACE and CCA-SSA. 
We also take the recently proposed generative model GraphMAE as baseline, because it bridges the gap between generative graph SSL models and GCL methods in terms of classification tasks.
Table~\ref{results:cla}
summarizes the classification results on all the datasets.
From the table,
we see that
CCA-SSG, GraphMAE and \algo\ lead other competitors and they almost tie. 
This shows that \algo\  achieves comparable performance with the state-of-the-art methods in the node classification task.
Further,
with the significant advantage in link prediction and attribute inference tasks,
we conclude that 
\algo,
a VGAE-based graph SSL method, 
can generate versatile node representations that can be widely used in various downstream tasks.


\subsection{Parameter Analysis (RQ4)}
We end this section with a sensitivity analysis on the key hyperparameters in \algo, i.e., the structure masking rate $\alpha_1$ and the feature masking rate $\alpha_2$.
Specifically,
we explore the stability of \algo\ w.r.t. the perturbation of $\alpha_1$ and $\alpha_2$.
We conduct experiments on the link prediction task
by varying these parameters from 0 to 0.5, 
and keeping others fixed.
Figure~\ref{fig:sensi} illustrates the AUC and AP scores of \algo-v3 under different $\alpha_1$ and $\alpha_2$ values 
on Cora and Citeseer.
From the figure,
we see that 
\algo-v3 can give very stable performance over a wide range of $\alpha_1$ and $\alpha_2$ values, 
as shown by the plateau in the figure.
This demonstrates the insensitivity of \algo\ w.r.t. these two hyper-parameters.

%% file: tex/conclusion.tex
\section{Conclusions}
\label{sec:conclusion}

We studied generative graph SSL in this paper
and proposed \algo,
which enhances the family of VGAE on graph representation learning.
Specifically,
\algo\ adopts the hierarchical variational framework in SIG-VAE 
and mainly focuses on feature reconstruction and structure/feature masking.
On the one hand,
\algo\ co-embeds both nodes and features in the encoder and
computes their embeddings by 
assuming they are independent and correlated, respectively.
After that,
feature embeddings that contain rich semantic information on features
are combined with node embeddings to provide more fine-grained information for feature reconstruction in the decoder.
On the other hand,
we injected the masking mechanism 
into \algo\ by adding an additional layer to the hierarchical variational framework.
We conducted extensive experiments 
to evaluate the performance of \algo.
The results show that 
\algo\ significantly outperforms other competitors in link prediction and attribute inference, and achieves comparable results with them in node classification.
This further verifies the power of generative graph SSL methods in graph representation learning.